\documentclass{zapiski}
\originfo{529}{}{}{2026}
\date{}

\usepackage[cp1251]{inputenc}
\usepackage[T2A]{fontenc}
\usepackage[russian,english]{babel}

\usepackage{times}
\usepackage{url}
\usepackage{latexsym}
\usepackage{graphicx}
\usepackage{amssymb}
\usepackage{amsmath}
\usepackage{hyperref}
\usepackage{booktabs}
\usepackage{tabularx}
\usepackage{caption}
\usepackage{cite}
\usepackage{multirow}
\usepackage{float}
\restylefloat{table}

\english

\begin{document}

\title[Data Filtering Methods for Training Language Models]{Data Filtering Methods for Training Language Models}

\author[E.~Shevchenko, E.~Bruches]{E.~Shevchenko and E.~Bruches}
\address[E.~Shevchenko]{Novosibirsk State University, Novosibirsk, Russia}
\email[E.~Shevchenko]{eggkolf@gmail.com}
\address[E.~Bruches]{Novosibirsk State University; A.~P.~Ershov Institute of Informatics Systems SB RAS, Novosibirsk, Russia}
\email[E.~Bruches]{bruches@bk.ru}


\begin{abstract}
Data quality is a critical factor in the effectiveness of machine learning models. Label errors, present even in widely used benchmarks, introduce noise into training data and reduce model generalization. In this work, we conduct a comparative analysis of two automatic label error detection methods~-- Confident Learning and Dataset Cartography~-- on three Russian text classification corpora of varying size, number of classes, and domain: ru\_emotion\_e-culture (49,123 examples, emotion classification), RuCoLA (8,524 examples, linguistic acceptability), and TERRa (2,337 examples, textual entailment recognition). We use the pre-trained rubert-base-cased model fine-tuned on each corpus. To verify the meaningfulness of filtering, we conduct control experiments with random removal of an equivalent number of examples. Results show that the effectiveness of both methods depends strongly on dataset characteristics: on large corpora with low noise levels, filtering does not improve performance, while on small datasets with high noise, Confident Learning achieves a significant F1-macro improvement. Dataset Cartography demonstrates more conservative behavior, removing fewer examples. Across all corpora, targeted removal by both methods outperforms random removal, confirming the meaningfulness of the approaches.
\end{abstract}

\maketitle

\section{Introduction}
\label{sec:intro}

The supervised learning paradigm assumes the availability of labeled data, the quality of which directly determines the upper bound on model effectiveness. In recent years, the research community's focus has shifted from developing increasingly complex architectures to improving training data quality~-- a direction known as \textit{data-centric AI}. Central to this direction is the problem of label noise, i.e., the discrepancy between the assigned label and the true category of an example.

The scale of the problem is significant. Northcutt et al.~\cite{northcutt2021confident} demonstrated that label errors are present even in widely used benchmarks: the authors found an average of 3.4\,\% erroneous labels in the test sets of ten popular datasets, including ImageNet, CIFAR-10, and Amazon Reviews. Correcting the discovered errors led to changes in model rankings on these benchmarks, calling into question the reliability of comparative evaluations based on noisy data. For training sets, which are typically less carefully verified than test sets, the error rate may be even higher.

For Russian-language corpora, this problem has been studied significantly less. Despite the emergence of large Russian-language benchmarks such as Russian SuperGLUE~\cite{shavrina2020superglue} and specialized corpora for individual NLP tasks, systematic analysis of annotation quality in these resources has hardly been conducted. Meanwhile, the characteristics of the Russian language~-- free word order, rich morphology, homonymy~-- potentially complicate the annotation task and increase the likelihood of annotator errors.

In this work, we conduct a comparative analysis of two automatic label error detection methods~-- Confident Learning~\cite{northcutt2021confident} and Dataset Cartography~\cite{swayamdipta2020cartography}~-- on three Russian-language corpora varying in size (from 2.3K to 49K training examples), number of classes (from 2 to 5), and domain (emotion classification, linguistic acceptability, textual entailment recognition). Both methods allow automatic identification of potentially mislabeled examples without manual verification, but rely on fundamentally different signals: Confident Learning uses out-of-fold model predictions obtained through cross-validation, while Dataset Cartography analyzes model training dynamics over several epochs.

It is important to emphasize that the two methods rely on fundamentally different signals for identifying problematic examples. Confident Learning uses \textit{consensus across models}: an example is considered erroneous if a model trained on independent data confidently predicts a different class for it. Dataset Cartography uses \textit{temporal training dynamics}: an example is considered problematic if the model consistently fails to learn it over several epochs. Comparing these two orthogonal approaches on the same data allows us to assess the extent to which error detection results depend on the chosen method and which approach is more effective under different conditions.

The contributions of this work are as follows:
\begin{enumerate}
    \item We conduct, to our knowledge, the first comparative analysis of Confident Learning and Dataset Cartography methods on Russian text classification corpora.
    \item We demonstrate a substantial dependence of filtering effectiveness on dataset characteristics~-- corpus size and the initial level of annotation noise.
    \item For each method, we conduct a control experiment with random removal of an equivalent subset, allowing us to separate the effect of targeted filtering from the effect of simple sample size reduction.
\end{enumerate}

\section{Related Work}
\label{sec:related}

\subsection{Confident Learning}

The Confident Learning method, proposed by Northcutt et al.~\cite{northcutt2021confident}, is based on estimating the joint distribution between observed (noisy) and latent (true) labels. The key concept of the method is the \textit{confident joint}~-- a matrix whose elements estimate the number of examples assigned a given label but whose true label differs. To construct this matrix, out-of-fold predicted probabilities are used, obtained through $K$-fold cross-validation: the model is trained on $K-1$ folds, and predictions are made for the remaining fold, which eliminates the effect of overfitting on the estimates.

An example is considered a potential label error if the model's predicted probability for an alternative class exceeds an adaptive threshold. This threshold is defined as the average predicted probability for the given class among all examples bearing that label. Thus, the method adapts to varying levels of model confidence across different classes.

The practical implementation of the method is available in the cleanlab library, which automates the process of label error detection and label quality metric computation. The method has demonstrated effectiveness across a wide range of tasks and data modalities, but its application to Russian-language text corpora has not been previously investigated.

\subsection{Dataset Cartography}

The Dataset Cartography method, proposed by Swayamdipta et al.~\cite{swayamdipta2020cartography}, takes a fundamentally different approach to data quality assessment. Instead of cross-validation, the method analyzes \textit{training dynamics}~-- changes in model predictions for each training example over several training epochs. For each example, four metrics are computed:

\begin{itemize}
    \item \textbf{Confidence}~-- the mean predicted probability of the true label, averaged across all training epochs.
    \item \textbf{Variability}~-- the standard deviation of this probability across epochs, reflecting prediction stability.
    \item \textbf{Correctness}~-- the fraction of epochs in which the model correctly classified the example.
    \item \textbf{Forgetfulness}~-- the number of transitions from ``correctly classified'' to ``incorrectly classified'' after the first successful learning.
\end{itemize}

Based on confidence and variability values, examples are divided into three regions: \textit{easy-to-learn}~-- high confidence and low variability; \textit{ambiguous}~-- high variability; \textit{hard-to-learn}~-- low confidence and low variability. Examples in the hard-to-learn region are the primary candidates for label errors: the model consistently predicts a class different from the given label.

A key advantage of the method is computational efficiency: unlike Confident Learning, which requires training $K$~models, Dataset Cartography requires training only one model (though for a greater number of epochs). A disadvantage is the use of in-sample predictions, which potentially leads to inflated confidence for examples on which the model has overfitted.

\subsection{Russian-Language Corpora and Quality Assessment}

The development of Russian-language resources for NLP has accelerated significantly with the emergence of the Russian SuperGLUE benchmark~\cite{shavrina2020superglue}, which includes nine tasks for evaluating Russian language understanding. The TERRa corpus used in this work is one of the tasks in this benchmark. The RuCoLA corpus~\cite{mikhailov2022rucola} was created by analogy with the English CoLA for evaluating linguistic acceptability of Russian sentences and includes error type annotations. The pre-trained rubert-base-cased model~\cite{kuratov2019rubert}, based on the BERT architecture and adapted for Russian, is a standard baseline for Russian text classification tasks.

Despite the active development of Russian-language corpora, systematic studies of annotation quality in them are practically absent. Existing works are typically limited to measuring inter-annotator agreement at the corpus creation stage but do not conduct post-hoc analysis of label errors using automatic methods. The present work aims to partially fill this gap by applying two well-established label error detection methods to three Russian-language corpora of varying scale and domain.

\section{Data}
\label{sec:data}

Experiments were conducted on three Russian-language corpora that differ substantially in size, number of classes, and task type. This selection allows investigating the behavior of filtering methods under different conditions. An overview of corpus characteristics is presented in Table~\ref{tab:datasets}.

\begin{table}[!tbh]
\centering
\caption{Characteristics of the corpora used.}
\label{tab:datasets}
\begin{tabular}{lccrrr}
\toprule
\textbf{Corpus} & \textbf{Task} & \textbf{Cls.} & \textbf{Train} & \textbf{Val} & \textbf{Test} \\
\midrule
ru\_emotion\_e-culture & Emotion & 5 & 49\,123 & 6\,140 & 6\,141 \\
RuCoLA & Acceptability & 2 & 8\,524 & 1\,066 & 1\,066 \\
TERRa & Entailment & 2 & 2\,337 & 292 & 293 \\
\bottomrule
\end{tabular}
\end{table}

\subsection{ru\_emotion\_e-culture}

The ru\_emotion\_e-culture corpus~\cite{eculture2026} is a collection of messages from a Russian-language forum, labeled with five emotional categories: \textit{aggression}, \textit{anxiety}, \textit{neutral}, \textit{positive}, and \textit{sarcasm}. The training set contains 49\,123 examples. This corpus is the largest of those used and is characterized by a relatively balanced class distribution. The specifics of the task~-- classifying short informal texts containing slang, profanity, and non-standard spelling~-- result in high variability of linguistic means and potential subjectivity of annotation, especially in distinguishing between the \textit{sarcasm} and \textit{neutral} categories.

\subsection{RuCoLA}

The RuCoLA (Russian Corpus of Linguistic Acceptability) corpus~\cite{mikhailov2022rucola} is designed for binary classification of linguistic acceptability of Russian sentences. Each example is labeled as \textit{acceptable} or \textit{unacceptable}, with unacceptable sentences annotated with error type: Semantics, Syntax, Morphology, or Hallucination. The training set contains 8\,524 examples. The class distribution is substantially imbalanced: the \textit{acceptable} class predominates. This task is of particular interest for annotation quality analysis, since the assessment of linguistic acceptability is subjective and depends on the annotator's linguistic competence.

\subsection{TERRa}

The TERRa (Textual Entailment Recognition for Russian) corpus~\cite{shavrina2020superglue} is part of the Russian SuperGLUE benchmark and represents the textual entailment recognition task. Each example consists of a pair of sentences~-- a premise and a hypothesis~-- and is labeled with one of two classes: \textit{entailment} or \textit{not\_entailment}. The training set contains 2\,337 examples, making TERRa the smallest of the three corpora. The textual entailment recognition task requires the model to understand semantic relations between sentences and is one of the most challenging NLU tasks. Input data is tokenized as a sentence pair: \texttt{[CLS] premise [SEP] hypothesis [SEP]}.

\subsection{Preprocessing and Data Splitting}

For all three corpora, uniform stratified splitting in an 80/10/10 ratio (train/validation/test) was applied with a fixed random seed to ensure reproducibility. Stratification by class labels guarantees preservation of class proportions across all subsets. Texts were tokenized using the rubert-base-cased~\cite{kuratov2019rubert} tokenizer with a maximum sequence length of 256 tokens. The validation and test sets remained unchanged throughout all experiments; filtering was applied exclusively to the training set.

\section{Methodology}
\label{sec:method}

\subsection{Base Model}

The pre-trained rubert-base-cased model~\cite{kuratov2019rubert}, developed by the DeepPavlov laboratory based on the BERT-base architecture~\cite{devlin2019bert}, was used as the base model. The model contains 12 transformer layers, 768 hidden units, and 12 attention heads, and was pre-trained on a Russian-language corpus using masked language modeling and next sentence prediction tasks.

Fine-tuning was performed using the Hugging Face Transformers library with the following hyperparameters: learning rate $2 \times 10^{-5}$ (linear schedule with warmup), batch size~32, number of epochs~3, maximum sequence length 256~tokens. Optimizer~-- AdamW. Training was conducted on GPU using mixed precision when compatible hardware was available. A fixed random seed was set for each experiment. The best model was selected based on F1-macro on the validation set, with final evaluation performed on the test set.

\subsection{Confident Learning}

The Confident Learning method was implemented using the cleanlab library~\cite{cleanlab2025}. The procedure consists of the following steps:

\begin{enumerate}
    \item \textbf{$K$-fold cross-validation} ($K = 4$). The training set is split into 4 folds. For each fold, the rubert-base-cased model is trained from scratch on the remaining three folds (3 epochs, same hyperparameters) and predicts class probabilities for examples in the held-out fold. As a result, each training example receives a vector of out-of-fold predicted probabilities $\hat{p}(y | x_i)$.

    \item \textbf{Label error detection}. Based on the predicted probability matrix and given labels, the \texttt{cleanlab.filter.find\_label\_issues()} function constructs the confident joint~-- a joint distribution matrix~-- and identifies examples with potential label errors. For each example, a \textit{label quality score} is computed~-- a numerical estimate of label reliability from 0 (likely error) to 1 (reliable label).

    \item \textbf{Filtering}. All examples identified as label errors are removed from the training set. The cleaned set is saved for subsequent retraining.
\end{enumerate}

\subsection{Dataset Cartography}

The Dataset Cartography method was implemented via a custom callback class for the Hugging Face Transformers library. The procedure consists of the following steps:

\begin{enumerate}
    \item \textbf{Collecting training dynamics}. The rubert-base-cased model is trained on the full training set for 10 epochs. The increased number of epochs (compared to 3 for the base model) is necessary to obtain statistically stable estimates of dynamic characteristics. After each epoch, the custom \texttt{DataMapsCallback} performs prediction on the entire training set and saves the logits for each example.

    \item \textbf{Computing metrics}. For each training example, four metrics are computed over 10 epochs: confidence (mean value of $p(y_i | x_i)$ across epochs), variability (standard deviation of $p(y_i | x_i)$), correctness (number of epochs with correct prediction), and forgetfulness (number of transitions from ``correct'' to ``incorrect'').

    \item \textbf{Example categorization}. Based on the computed metrics, each example is assigned to one of three categories. Median values of confidence and variability are used as thresholds: examples with variability above the median are classified as \textit{ambiguous}; of the remaining~-- examples with confidence above the median are classified as \textit{easy-to-learn}, the rest as \textit{hard-to-learn}.

    \item \textbf{Filtering}. An automatic threshold based on the knee-point detection method is applied for example removal: examples are sorted by confidence, the point of maximum curvature is determined, and all examples with confidence below the threshold are removed from the training set.
\end{enumerate}

\subsection{Control Experiment}

For each filtering method, a control experiment was conducted with random removal of a subset from the training set equal in size to the number of examples removed by the corresponding method. Random removal was performed with a fixed seed for reproducibility. This experiment allows separating the effect of targeted filtering from the effect of simple training set size reduction. If the filtering method significantly outperforms random removal, this indicates the meaningfulness of the identified examples.

\subsection{Evaluation Metric}

\textbf{F1-macro}~-- the arithmetic mean of F1-scores computed independently for all classes~-- was used as the primary metric. This metric equally weights all classes regardless of their size and is therefore robust to class imbalance present in the RuCoLA and TERRa corpora. Accuracy was additionally recorded.

\section{Results}
\label{sec:results}

\subsection{Filtering Statistics}

The two methods demonstrate substantially different degrees of filtering aggressiveness on the same data (Table~\ref{tab:filtering}). Hereafter, Confident Learning is abbreviated as CL and Dataset Cartography as DM.

\begin{table}[!tbh]
\centering
\caption{Training set filtering statistics.}
\label{tab:filtering}
\begin{tabular}{llrrr}
\toprule
\textbf{Corpus} & \textbf{Method} & \textbf{Removed} & \textbf{\%} & \textbf{Remaining} \\
\midrule
\multirow{2}{*}{ru\_emotion\_e-culture} & CL & 1\,841 & 3.75 & 47\,282 \\
 & DM & 4\,558 & 9.28 & 44\,565 \\
\midrule
\multirow{2}{*}{RuCoLA} & CL & 1\,567 & 18.38 & 6\,957 \\
 & DM & 1\,234 & 14.48 & 7\,290 \\
\midrule
\multirow{2}{*}{TERRa} & CL & 829 & 35.47 & 1\,508 \\
 & DM & 256 & 10.95 & 2\,081 \\
\bottomrule
\end{tabular}
\end{table}

A notable pattern emerges: the proportion of examples identified by Confident Learning increases as corpus size decreases~-- from 3.75\,\% for ru\_emotion\_e-culture to 35.47\,\% for TERRa. Dataset Cartography demonstrates more stable behavior: the proportion of removed examples ranges from 9.28\,\% to 14.48\,\%.

On the ru\_emotion\_e-culture corpus, Dataset Cartography removes 2.5 times more examples than Confident Learning (4,558 vs. 1,841), while on TERRa the ratio is inverted: Confident Learning removes 3.2 times more (829 vs. 256). This difference reflects the fundamentally different mechanisms of problematic example identification employed by the two methods.

\subsection{Classification Results}

F1-macro values on the test set for different training conditions are presented in Table~\ref{tab:f1results}; differences between methods are shown in Table~\ref{tab:deltas}.

\begin{table}[!tbh]
\centering
\caption{F1-macro on the test set.}
\label{tab:f1results}
\begin{tabular}{lccccc}
\toprule
\textbf{Corpus} & \textbf{Base} & \textbf{CL} & \textbf{DM} & \textbf{Rnd(CL)} & \textbf{Rnd(DM)} \\
\midrule
e-culture & 0.9353 & 0.9320 & 0.9260 & 0.9325 & 0.9237 \\
RuCoLA & 0.6635 & 0.6262 & 0.6438 & 0.6096 & 0.6297 \\
TERRa & 0.6444 & 0.6578 & 0.6441 & 0.5230 & 0.5851 \\
\bottomrule
\end{tabular}
\end{table}

\begin{table}[!tbh]
\centering
\caption{F1-macro differences ($\Delta$) between methods.}
\label{tab:deltas}
\begin{tabular}{lcccc}
\toprule
\textbf{Corpus} & \textbf{CL--Base} & \textbf{DM--Base} & \textbf{CL--Rnd} & \textbf{DM--Rnd} \\
\midrule
e-culture & $-$0.0033 & $-$0.0093 & $-$0.0005 & +0.0023 \\
RuCoLA & $-$0.0373 & $-$0.0197 & +0.0166 & +0.0141 \\
TERRa & +0.0134 & $-$0.0003 & +0.1348 & +0.0590 \\
\bottomrule
\end{tabular}
\end{table}

\subsection{ru\_emotion\_e-culture}

On the largest of the three corpora, neither filtering method led to improved classification quality. Confident Learning identified 1,841 potential label errors (3.75\,\% of the training set), predominantly in the \textit{sarcasm} (535 examples) and \textit{neutral} (529 examples) classes, which is explained by the subjectivity of distinguishing these categories. After removing the identified examples, F1-macro was 0.9320~-- 0.0033 below the baseline (0.9353). The control experiment with random removal of 1,841 examples yielded F1-macro\,=\,0.9325, which is virtually identical to the Confident Learning result (a difference of only 0.0005).

Dataset Cartography filtered 4,558 examples (9.28\,\%) and achieved F1-macro\,=\,0.9260, which is 0.0093 below the baseline. The more aggressive filtering led to a more noticeable, though still small, decrease in quality.

These results indicate that on a large corpus with initially high annotation quality (baseline F1-macro\,>\,0.93), the model is able to independently mitigate the influence of sparse label errors, and their removal does not yield a noticeable improvement.

\subsection{RuCoLA}

On the medium-sized corpus, both methods identified a substantial proportion of potential errors. Confident Learning identified 1,567 examples (18.38\,\%), with the vast majority (1,289 of 1,567) belonging to the \textit{unacceptable} class. The distribution by error type shows that the greatest number of problematic annotations are associated with semantic (615) and syntactic (505) errors. Dataset Cartography removed 1,234 examples (14.48\,\%), also predominantly from the \textit{unacceptable} class (1,160 of 1,234).

Neither method improved absolute classification quality: F1-macro after Confident Learning was 0.6262 ($-$0.0373 relative to baseline), after Dataset Cartography~-- 0.6438 ($-$0.0197). However, both methods substantially outperformed random removal: Confident Learning showed a result 0.0166 higher than random removal of equivalent size (0.6262 vs. 0.6096), Dataset Cartography~-- 0.0141 higher (0.6438 vs. 0.6297). This confirms that both methods identify substantively more problematic examples than a random subsample.

It is noteworthy that Dataset Cartography, while removing fewer examples (1,234 vs. 1,567), demonstrates a less pronounced F1-macro decrease ($-$0.0197 vs. $-$0.0373), suggesting more targeted filtering.

\subsection{TERRa}

The smallest of the three corpora yields the most expressive results. Confident Learning identified 829 potential errors, constituting 35.47\,\% of the training set~-- an exceptionally high figure indicating a significant level of annotation noise in this corpus. Of the 829 identified examples, 470 belong to the \textit{not\_entailment} class and 359 to the \textit{entailment} class.

After removing the identified errors and retraining the model, F1-macro increased from 0.6444 to 0.6578 (+0.0134)~-- this is the only case across all experiments where filtering led to an improvement in absolute quality, despite reducing the training set by more than a third (from 2,337 to 1,508 examples). The control experiment with random removal of 829 examples showed a sharp drop in F1-macro to 0.5230 ($-$0.1214 relative to baseline), and the difference between Confident Learning and random removal was +0.1348~-- the largest across all experiments.

Dataset Cartography on the same corpus removed only 256 examples (10.95\,\%) and achieved F1-macro\,=\,0.6441~-- virtually at the baseline level ($-$0.0003). Meanwhile, random removal of 256 examples led to a drop to 0.5851 ($-$0.0593 relative to baseline), and the difference between Dataset Cartography and random removal was +0.0590.

The difference in behavior of the two methods on this corpus is the most illustrative. Confident Learning, by identifying and removing a significantly larger proportion of noisy examples, achieves a real improvement in quality. Dataset Cartography, acting more conservatively, maintains quality at the baseline level but does not provide an improvement.

\subsection{Summary of Results}

Summary results of all experiments are presented in Table~\ref{tab:summary}.

\begin{table}[!tbh]
\centering
\caption{Summary results: filtering statistics and F1-macro on the test set.}
\label{tab:summary}
\begin{tabular}{llrrcrr}
\toprule
\textbf{Corpus} & \textbf{Method} & \textbf{Rem.} & \textbf{\%} & \textbf{F1} & \textbf{$\Delta$Base} & \textbf{$\Delta$Rnd} \\
\midrule
\multirow{3}{*}{e-culture}
 & Baseline & -- & -- & 0.9353 & -- & -- \\
 & CL & 1\,841 & 3.75 & 0.9320 & $-$0.0033 & $-$0.0005 \\
 & DM & 4\,558 & 9.28 & 0.9260 & $-$0.0093 & +0.0023 \\
\midrule
\multirow{3}{*}{RuCoLA}
 & Baseline & -- & -- & 0.6635 & -- & -- \\
 & CL & 1\,567 & 18.38 & 0.6262 & $-$0.0373 & +0.0166 \\
 & DM & 1\,234 & 14.48 & 0.6438 & $-$0.0197 & +0.0141 \\
\midrule
\multirow{3}{*}{TERRa}
 & Baseline & -- & -- & 0.6444 & -- & -- \\
 & CL & 829 & 35.47 & 0.6578 & +0.0134 & +0.1348 \\
 & DM & 256 & 10.95 & 0.6441 & $-$0.0003 & +0.0590 \\
\bottomrule
\end{tabular}
\end{table}

\section{Analysis and Discussion}
\label{sec:discussion}

\subsection{Effect of Corpus Size and Noise Level}

The obtained results demonstrate a clear dependence of filtering effectiveness on two interrelated corpus characteristics: the size of the training set and the level of annotation noise.

On the large ru\_emotion\_e-culture corpus (49,123 examples), characterized by high baseline quality (F1-macro\,=\,0.9353), both methods identify a small proportion of potential errors (3.75\,\% for CL, 9.28\,\% for DM), and their removal does not improve performance. This is consistent with the results of Northcutt et al.~\cite{northcutt2021confident}, who showed that deep learning models possess a certain robustness to label noise, especially when the error rate is low.

On the small TERRa corpus (2,337 examples), Confident Learning identifies an anomalously high proportion of potential errors (35.47\,\%). Although some of these may be false positive identifications caused by insufficient data for training within cross-validation folds, the improvement in F1-macro after filtering (+0.0134) and the dramatic superiority over random removal (+0.1348) convincingly confirm the presence of significant real noise in this corpus's annotations.

The RuCoLA corpus occupies an intermediate position: with a medium size (8,524 examples) and a moderate proportion of identified errors (14--18\,\%), filtering does not improve absolute quality, but targeted removal consistently outperforms random removal.

\subsection{Comparison of Filtering Strategies}

Confident Learning and Dataset Cartography demonstrate fundamentally different filtering strategies. Confident Learning relies on out-of-fold predictions and identifies examples for which a model trained on independent data predicts a class different from the given label. This approach is more aggressive on small datasets (35.47\,\% on TERRa) and less aggressive on large ones (3.75\,\% on e-culture).

Dataset Cartography analyzes the dynamics of in-sample predictions and categorizes examples based on prediction stability over time. The method is more stable in terms of the proportion of removed examples (9--14\,\% across all corpora) and generally acts more conservatively. This makes Dataset Cartography more predictable and safer: the method is less likely to remove correctly labeled examples, but may miss real label errors.

An important practical characteristic is the computational cost of the methods. Confident Learning requires training $K = 4$ models within cross-validation, which increases computational costs by approximately 4 times compared to a single training run. Dataset Cartography trains one model but for an increased number of epochs, which only slightly increases costs. Thus, Dataset Cartography is more computationally efficient, which may be critical for large corpora.

The difference in filtering aggressiveness has direct practical implications. On the TERRa corpus, Confident Learning removes more than a third of the training set, which on one hand improves quality, but on the other~-- creates a risk of excessive data reduction. Dataset Cartography, removing only 11\,\%, minimizes this risk, albeit at the cost of no quality improvement.

\subsection{Meaningfulness of Filtering}

A key result of this work is the demonstration of the meaningfulness of both filtering methods. Across all corpora for which a control experiment was conducted, targeted removal of examples by both methods outperforms random removal of equivalent size. This is most convincingly demonstrated on the TERRa corpus: Confident Learning outperforms random removal by 0.1348 in F1-macro, Dataset Cartography~-- by 0.0590. Even on the RuCoLA corpus, where absolute quality decreases after filtering, both methods show results above random removal (by 0.0166 for CL and 0.0141 for DM).

This result is non-trivial: it indicates that the methods genuinely identify examples with lower annotation quality, and the observed effects cannot be explained by simple reduction of training set size.

It should be acknowledged, however, that the comparison with random removal is an indirect measure of filtering quality: it confirms that the methods do not act as random subsampling, but does not establish their precision and recall against a ground-truth set of true label errors. A direct assessment of these characteristics would require manual expert verification of a sample of identified examples, which we leave for future work (see Section~\ref{sec:future}).

\subsection{Characteristics of Identified Examples}

Analysis of the distribution of identified errors by class allows identifying the most problematic categories. On the ru\_emotion\_e-culture corpus, the greatest number of errors was found in the \textit{sarcasm} (535, CL) and \textit{neutral} (529, CL) classes, which is explained by the subjectivity of distinguishing these categories. On the RuCoLA corpus, the vast majority of errors fall in the \textit{unacceptable} class (1,289 of 1,567 for CL; 1,160 of 1,234 for DM), indicating higher uncertainty in annotating unacceptable sentences. By error type, semantic and syntactic violations lead, pointing to the difficulty of their unambiguous identification by annotators.

Dataset Cartography provides an additional characterization of the corpus as a whole. The model's average confidence on ru\_emotion\_e-culture (0.979) substantially exceeds confidence on TERRa (0.811), which is consistent with the difference in baseline model quality on these corpora. Average variability on TERRa (0.184) is more than five times the value on ru\_emotion\_e-culture (0.033), reflecting substantially greater annotation ambiguity.

\subsection{Training Dynamics Statistics}

The Dataset Cartography method provides additional diagnostic information about corpus quality in the form of aggregated training dynamics statistics (Table~\ref{tab:datamaps_stats}).

\begin{table}[!tbh]
\centering
\caption{Dataset Cartography statistics by corpus.}
\label{tab:datamaps_stats}
\begin{tabular}{lccccrrr}
\toprule
\textbf{Corpus} & $\bar{\mu}$ & $\bar{\sigma}$ & \textbf{Corr.} & \textbf{Thresh.} & \textbf{Easy} & \textbf{Amb.} & \textbf{Hard} \\
\midrule
e-culture & 0.979 & 0.033 & 6.90 & 0.940 & 23\,929 & 24\,561 & 633 \\
RuCoLA & 0.912 & 0.104 & 6.60 & 0.834 & 3\,906 & 4\,262 & 356 \\
TERRa & 0.811 & 0.184 & 6.19 & 0.734 & 739 & 1\,168 & 430 \\
\bottomrule
\end{tabular}
\end{table}

The statistics demonstrate a clear correlation with baseline classification quality. The ru\_emotion\_e-culture corpus, on which the model achieves the highest F1-macro (0.9353), is characterized by high average confidence $\bar{\mu}$ (0.979) and minimal variability $\bar{\sigma}$ (0.033), indicating stable and confident predictions for the vast majority of examples. The proportion of hard-to-learn examples is only 1.3\,\% (633 of 49,123).

At the opposite end of the spectrum is the TERRa corpus, characterized by significantly lower confidence $\bar{\mu}$ (0.811), high variability $\bar{\sigma}$ (0.184), and the largest proportion of hard-to-learn examples~-- 18.4\,\% (430 of 2,337). This is consistent with the high difficulty of the textual entailment recognition task and significant annotation noise.

The RuCoLA corpus occupies an intermediate position across all metrics. Notably, average forgetfulness on the TERRa corpus (0.071) is more than 3.5 times the value for ru\_emotion\_e-culture (0.021), further confirming training instability on this corpus.

Thus, Dataset Cartography statistics can be used as a standalone diagnostic tool for corpus quality, allowing assessment of annotation noise levels prior to filtering.

\subsection{Limitations}

This study has several limitations that should be taken into account when interpreting the reported results.

First, no manual expert verification of identified examples was conducted. As a consequence, the precision and recall of the methods against a ground-truth set of true label errors remain unknown, and the meaningfulness of filtering is established only indirectly~-- through comparison with random removal of an equivalent number of examples.

Second, all experiments were conducted using a single base model (rubert-base-cased). The training dynamics observed for Dataset Cartography and the cross-validated probabilities used by Confident Learning are, in part, properties of this specific architecture, and the extent to which our quantitative findings generalize to other Russian-language encoders or to multilingual models is not directly addressed by our experimental design.

Third, each experiment was performed with a single fixed random seed. The reported differences in F1-macro between filtering, baseline, and random removal conditions therefore reflect single-run point estimates rather than averages over multiple runs, and we do not report measures of variance or formal tests of statistical significance. The qualitative pattern across corpora (no improvement on the large clean corpus, intermediate behavior on RuCoLA, clear improvement of Confident Learning on TERRa) is consistent with the substantial differences in dataset size and noise level, but the magnitude of individual numerical differences should be interpreted with corresponding caution.

Finally, the analysis is restricted to text classification. Generalization to other NLP tasks~-- such as sequence labeling, generation, or information extraction~-- requires additional study, since the notion of a ``label error'' and the applicability of out-of-fold predictions and per-example training dynamics differ for these settings.

\section{Future Work}
\label{sec:future}

The limitations identified above directly motivate several directions for future work. The most immediate is manual expert verification of a sample of examples flagged by Confident Learning and Dataset Cartography, which would allow direct estimation of precision and recall and a more rigorous comparison of the two methods beyond their effect on downstream F1-macro. A complementary direction is to repeat the experiments across multiple random seeds and additional Russian-language encoders (e.g., other rubert variants, ruRoBERTa, multilingual XLM-R) in order to disentangle method-level effects from architecture-specific or run-specific variation and to support formal tests of statistical significance.

In addition, we plan to develop a hybrid method combining the statistical confidence of Confident Learning with the training dynamics analysis of Dataset Cartography. The two methods rely on orthogonal signals~-- out-of-fold predictions and temporal training dynamics, respectively~-- which creates prerequisites for improving the reliability of error detection through consensus: examples identified by both methods simultaneously are highly likely to be genuine label errors. A further promising direction is the evaluation of LLM-based methods for data quality assessment, in particular the DS2 approach~\cite{pang2025ds2}.

\section{Conclusion}
\label{sec:conclusion}

In this work, we conducted a comparative analysis of two automatic label error detection methods~-- Confident Learning and Dataset Cartography~-- on three Russian text classification corpora. The main results are as follows:

\begin{enumerate}
    \item \textbf{Dependence on dataset characteristics.} The effectiveness of both methods depends substantially on corpus size and the initial level of annotation noise. On large corpora with a low error rate (less than 4\,\%), filtering does not improve classification quality. On small corpora with high noise (TERRa, 35.5\,\% identified errors), Confident Learning achieves a significant F1-macro improvement (+0.0134).

    \item \textbf{Difference in strategies.} Confident Learning demonstrates adaptive behavior, varying the proportion of removed examples from 3.75\,\% to 35.47\,\% depending on the corpus. Dataset Cartography acts more conservatively and stably (9--14\,\% across all corpora), making it more predictable but less effective under high noise levels.

    \item \textbf{Meaningfulness of filtering.} Across all corpora, targeted removal of examples by both methods outperforms random sample reduction of equivalent size. The greatest superiority is observed on the TERRa corpus: +0.1348 for Confident Learning and +0.0590 for Dataset Cartography compared to random removal.

    \item \textbf{Practical recommendation.} For small Russian-language corpora with presumed high noise levels, the application of Confident Learning with subsequent verification and retraining is recommended. For medium-sized corpora, Dataset Cartography can be used as a safer tool for data quality diagnostics.
\end{enumerate}

These findings should be read jointly with the limitations discussed above: the reported numerical differences correspond to single-seed runs of a single base model and have not been validated against manually verified label errors, so the conclusions are most robust at the qualitative level (relative ordering of methods and dependence on corpus characteristics) rather than at the level of individual F1-macro values.

\section*{Acknowledgments}

The author thanks Elena Bruches for scientific supervision and valuable discussions throughout this work.

\providecommand{\bysame}{\leavevmode\hbox to3em{\hrulefill}\thinspace}
\providecommand{\MR}{\relax\ifhmode\unskip\space\fi MR }
\providecommand{\MRhref}[2]{\href{http://www.ams.org/mathscinet-getitem?mr=#1}{#2}}
\providecommand{\href}[2]{#2}

\bigskip

\selectlanguage{russian}

\noindent
Е.~Шевченко, Е.~Бручес. \emph{Методы фильтрации данных для обучения языковых моделей.}

\medskip

\noindent
\textbf{Аннотация.} Качество данных является критическим фактором эффективности моделей машинного обучения. Ошибки разметки, присутствующие даже в широко используемых эталонных наборах, вносят шум в обучающие данные и снижают обобщающую способность моделей. В настоящей работе проводится сравнительный анализ двух автоматических методов обнаружения ошибок разметки~--- Confident Learning и Dataset Cartography~--- на трёх русскоязычных корпусах текстовой классификации, различающихся по объёму, числу классов и предметной области: ru\_emotion\_e-culture (49\,123 примера, классификация эмоций), RuCoLA (8\,524 примера, лингвистическая приемлемость) и TERRa (2\,337 примеров, распознавание текстовой импликации). В качестве базовой модели используется предобученная rubert-base-cased, дообучаемая на каждом корпусе. Для проверки осмысленности фильтрации проводится контрольный эксперимент со случайным удалением эквивалентного числа примеров. Результаты показывают, что эффективность обоих методов существенно зависит от характеристик набора данных: на больших корпусах с низким уровнем шума фильтрация не улучшает качество, тогда как на малых корпусах с высоким уровнем шума Confident Learning достигает заметного прироста F1-macro. Dataset Cartography демонстрирует более консервативное поведение, удаляя меньшее число примеров. На всех корпусах целенаправленное удаление обоими методами превосходит случайное удаление, что подтверждает осмысленность подхода.

\medskip

\noindent
\textbf{Ключевые слова:} качество данных, шум разметки, confident learning, dataset cartography, классификация текстов, обработка естественного языка.

\selectlanguage{english}

\end{document}